\documentclass[conference]{IEEEtran}
\IEEEoverridecommandlockouts
\usepackage{cite}
\usepackage{amsmath,amssymb,amsfonts}
\usepackage{algorithmic}
\usepackage{graphicx}
\usepackage{textcomp}

\usepackage{caption}
\usepackage{subcaption}
\usepackage{xcolor}
\usepackage{bm}
\newtheorem{remark}{Remark}
\usepackage{hyperref}

\def\BibTeX{{\rm B\kern-.05em{\sc i\kern-.025em b}\kern-.08em
    T\kern-.1667em\lower.7ex\hbox{E}\kern-.125emX}}
\begin{document}

\title{Availability Adversarial Attack and Countermeasures for Deep Learning-based Load Forecasting
}

\author{\IEEEauthorblockN{Wangkun Xu}
\IEEEauthorblockA{\textit{Electrical and Electronic Engineering} \\
\textit{Imperial College London}\\
London, UK \\
wangkun.xu18@imperial.ac.uk}
\and
\IEEEauthorblockN{Fei Teng}
\IEEEauthorblockA{\textit{Electrical and Electronic Engineering} \\
\textit{Imperial College London}\\
London, UK \\
f.teng@imperial.ac.uk}
}

\maketitle

\begin{abstract}
The forecast of electrical loads is essential for the planning and operation of the power system. Recently, advances in deep learning have enabled more accurate forecasts. However, deep neural networks are prone to adversarial attacks. Although most of the literature focuses on integrity-based attacks, this paper proposes availability-based adversarial attacks, which can be more easily implemented by attackers. For each forecast instance, the availability attack position is optimally solved by mixed-integer reformulation of the artificial neural network. To tackle this attack, an adversarial training algorithm is proposed. In simulation, a realistic load forecasting dataset is considered and the attack performance is compared to the integrity-based attack. Meanwhile, the adversarial training algorithm is shown to significantly improve robustness against availability attacks. All codes are available at \url{https://github.com/xuwkk/AAA_Load_Forecast}.
\end{abstract}


\section{Introduction}

\subsection{Load Forecasting}

Load forecasting plays an essential role in the operation and control of the power system. With more distributed resources embedded in the grid, accurate load forecasting also benefits the demand-side response. Plenty of researches have been done for load forecasting, both deterministic and probabilistic \cite{hong2016probabilistic}. Regression-based models and their variants dominates the long-term forecasting while machine learning algorithms, such as artificial neural network (ANN) gains more attention for short-term forecasting due to its efficiency on feature extraction \cite{kuster2017electrical}. For instance, feedforward neural network is applied in \cite{zhang2022cost}. In addition, long short-term memory (LSTM) and convolutional neural net (CNN) are used to remember temporal information, which can result in state-of-the-art accuracy \cite{rafi2021short}. 

\subsection{Adversarial Attack}

The concept of adversarial attack against deep neural network was firstly discovered by Szegedy \textit{et al.} \cite{szegedy2013intriguing} where it has been shown that a small amount of perturbation on the image pixels can result in wrong classification. To generate the attack, fast gradient sign method (FGSM) was proposed by assuming the local linearity of the loss surface \cite{goodfellow2014explaining}. Although the FGSM can reduce the computational burden, it can only produces sub-optimal attack vectors. Projected gradient descent (PGD) extends the idea of FGSM in which the attack vector is solved in a multistep manner to approach the global optima \cite{madry2017towards}. In addition, the most effective approach to combating adversarial attacks is to minimize the loss of adversarial examples generated at each epoch, which is named adversarial training \cite{madry2017towards}.

Recently, adversarial attacks on time series, especially the load forecasting applications, have gained more attention \cite{fawaz2019adversarial}. Starting from the linear model, Luo \textit{et al.} compares four load forecasting algorithms that can easily fail under data integrity attacks \cite{luo2018benchmarking}. The authors then introduce regularization to address the challenge \cite{luo2018robust}. However, the integrity attacks considered in \cite{luo2018benchmarking, luo2018robust} are not specifically designed by the forecasting model. In \cite{stratigakos2022towards}, a robust linear regression for data availability attack is proposed through min-max optimization. In the deep learning era, a cost-oriented integrity attack is established to maximize operational cost \cite{chen2022vulnerability}. In \cite{zhou2022robust}, the robustness of the deep load forecasting model is guaranteed through Bayesian inference.

\subsection{Contribution}

We observe that the integrity attack is the main source of adversarial attacks in deep learning-based load forecasting models, and the availability attack is only considered for linear models. Considering a `man-in-the-middle' scenario, availability attack is much cheaper than integrity attack as the attacker does not need to manipulate the data and repackage the packets \cite{pan2018cyber}. Meanwhile, missing data are more common, as measurements can be temporally unavailable due to equipment failure. To fill this research gap, first, we design the availability adversarial attack targeting on piecewise linear neural network (PLNN). The optimality of the attack is guaranteed by reformulating the PLNN into mixed integer linear programming (MILP). Second, an adversarial training algorithm is proposed that considers both clean and attack accuracy to improve the robustness of the PLNN. In the simulation, we compare the performance of the availability adversarial attack with its integrity counterpart and demonstrate the effectiveness of the proposed adversarial training algorithm in a realistic load forecasting task.

\section{Load Forecasting Model}

Load forecasting problem is defined as a supervised learning task where the input to the model is a collection of features, such as temperature, and the output is the true load consumption. Define a dataset space $\mathcal{X}\times\mathcal{Y}$ where $\mathcal{X}$ represents the input space and $\mathcal{Y}$ represents the target load space. To train a load forecasting model, a dataset $\bm{X}\times\bm{Y}$ can be sample from $\mathcal{X}\times\mathcal{Y}$ where $\bm{X} = [\bm{x}_1,\bm{x}_2,\cdots,\bm{x}_N]\in\mathbb{R}^{N\times p}$ and $\bm{Y} = [{y}_1,{y}_2,\cdots,{y}_N]\in\mathbb{R}^{N}$. In this paper, the dataset from \cite{farrokhabadi2022day} is used which consists of data from the four-year metropolitan load at a resolution of one hour. The forecast features include K1: air pressure (kW), K2: cloud cover in \%, K3: humidity in \%, K4: temperature in $^\circ C$, K5: wind direction in degrees, k6: wind speed in km/h, and K7: temporal information. Temporal information that includes the month, date, and time is further converted into sine and cosine values on the basis of different periods. Consequently, the above setting results in $p=12$ number of features.

A deep neural network $f(\cdot;\bm{\theta}): \mathcal{X}\rightarrow\mathcal{Y}$ parameterized on $\bm{\theta}$ can be trained to minimize the mean squared error (MSE) loss between the targeted and predicted loads:
\begin{equation}\label{eq:loss_function}
    \mathcal{L}(\bm{\theta}) = \frac{1}{N}\sum_{n=1}^{N}\left(y_n-f(\bm{x}_n;\bm{\theta})\right)^2
\end{equation}

Stochastic gradient descent (SGD) is broadly used to find the local minima $\bm{\theta}^*$ of \eqref{eq:loss_function}. At each epoch, the gradient is found in a batch $\bm{B}_n$ of the dataset:
\begin{equation}\label{eq:sgd_plain}
    \bm{\theta}:=\bm{\theta} - \frac{\alpha}{|\bm{B}_n|} \nabla_{\bm{\theta}}\left(\sum_{(\bm{x},\bm{y})\in\bm{B}_n}(y-f(\bm{x};\bm{\theta}))^2\right)
\end{equation}
where $\alpha$ is a user-defined learning rate.

Specifically, this paper considers a piecewise linear neural network structure with ReLU activation functions before the last layer. Therefore, the neural network $f(\bm{x};\bm{\theta})$ with $d+1$ layers ($d\geq2$) can be explicitly written as:
\begin{equation}\label{eq:nn_explicit}
\begin{split}
    \bm{z}_1 & = \bm{x} \\
    \bm{z}_{i+1} & = \max{\{\bm{0}, \bm{W}_i\bm{z}_i + \bm{b}_i \}},\quad i = 1,\cdots,d-1 \\
    \bm{z}_{d+1} & = \bm{W}_d \bm{z}_d + \bm{b}_d
\end{split}
\end{equation}
where $\bm{W}_i$ and $\bm{b}_i$ are the weight and bias of the $i$-th layer. 

\section{Adversarial Attacks on Load Forecasting}

In the literature, adversarial attack commonly refers to maliciously perturbing a small value in the input data $\bm{x}_n$, that is, integrity adversarial attack. The `adversarial' implies that the attacker can maliciously design the attack vector for a specific trained ANN and data instance under a white-box assumption. However, in this paper, we propose availability adversarial attack where the attacker can block part or all of the input features. 

\subsection{Integrity Adversarial Attacks}

Referring to the loss function \eqref{eq:loss_function}, the objective of the adversarial attack is to increase the forecast error by perturbing $\bm{\delta}$ on the input $\bm{x}_n$ subject to a predefined $l_p$ ball $\|\cdot\|_p$. After obtaining the trained model $\bm{\theta}^*$, the integrity adversarial attack on $(\bm{x}_n,y_n)$ can be written as
\begin{equation}\label{eq:loss_integrity}
    \max_{\|\bm{\delta}\|_p\leq \epsilon}(y_n - f(\bm{x}_n+\bm{\delta};\bm{\theta}^*))^2
\end{equation}

Projected gradient descent (PGD)\footnote{Though the objective of \eqref{eq:loss_integrity} is to maximize the loss; it is common to name it descent in the literature.} can be used to solve \eqref{eq:loss_integrity}. At each iteration, normalized gradient ascents are operated and the resulting data $\bm{x}_n + \bm{\delta}$ is clamped by the norm ball \cite{madry2017towards}.

Although PGD has proven to be efficient, it is hard to obtain the global optima due to the non-convexity of the neural network. To approach the global optimality, multi-run strategy can be applied. However, it is possible to find global optima for small sized neural network under certain structure. Firstly, since the square loss in \eqref{eq:loss_integrity} is convex, to formulate a convex problem, we can maximize and minimize the forecast values. Second, $l_\infty$-norm is considered in which each feature can be maximally injected by $\epsilon$. Referring to \eqref{eq:nn_explicit}, \eqref{eq:loss_integrity} can be rewritten as:
\begin{equation}\label{eq:inte_1}
\begin{split}
& {\min / \max}_{\bm{z}_{1},\dots,\bm{z}_{d+1}}  \bm{z}_{d+1}  \\
\text{s.t.} \quad & \|\bm{z}_1 - \bm{x}\|_\infty \leq \epsilon \\
& \bm{z}_{i+1} = \max{\{\bm{0}, \bm{W}_i\bm{z}_i + \bm{b}_i \}},\quad i = 1,\cdots,d-1 \\
& \bm{z}_{d+1} = \bm{W}_d \bm{z}_d + \bm{b}_d
\end{split}
\end{equation}

The optimization problem \eqref{eq:inte_1} is non-convex due to the maximization constraint caused by ReLU. However, it can be easily transformed into an MILP using the big-M method. Specifically, \eqref{eq:inte_1} is equivalent to \cite{liu2020certified}:
\begin{equation}\label{eq:inte_2}
\begin{split}
& {\min / \max}_{\bm{z}_{1},\dots, \bm{z}_{d+1}, \bm{v}_1, \ldots, \bm{v}_{d-1}}  \bm{z}_{d+1} \\
\text{s.t.} \quad  & \bm{z}_{i+1} \geq \bm{W}_i \bm{z}_i + \bm{b}_i, \quad i=1,\dots,d-1 \\
& \bm{z}_{i+1} \geq \bm{0}, \quad i=1, \dots, d-1 \\
& \bm{u}_i \cdot \bm{v}_i \geq  \bm{z}_{i+1}, \quad i=1, \dots,d-1 \\
& \bm{W}_i \bm{z}_i + \bm{b}_i \geq \bm{z}_{i+1} + (\bm{1}-\bm{v}_i) \bm{l}_i, \quad i=1,\ldots,d-1 \\
& \bm{v}_i \in \{0,1\}^{|\bm{v}_i|}, \quad i=1, \ldots,d-1 \\
& \bm{z}_1 \leq \bm{x} + \bm{\delta} \\
& \bm{z}_1 \geq \bm{x} - \bm{\delta} \\
& \bm{z}_{d+1} = \bm{W}_d \bm{z}_d + \bm{b}_d
\end{split}
\end{equation}
where $\bm{v}_i$ is the auxiliary integer variable related to the ReLU activation functions; $\bm{u}_i$ and $\bm{l}_i$ are the upper and lower bounds for the output of the $i$-th layer, i.e., $\bm{l}_i \leq \bm{W}_i\bm{z}_i + \bm{b}_i \leq \bm{u}_i, i=1,\cdots, d-1$, which are designed by the user. All the inequality constraints in \eqref{eq:inte_2} are element-wise. If the inequality is satisfied for some optimal solution $\bm{z}_i = \bm{z}_i^*, i=1,\cdots,d-1$ of \eqref{eq:inte_1}, then the optimal solution of \eqref{eq:inte_2} is also an optimal solution of \eqref{eq:inte_1} \cite{teixeira2015secure}. A detailed conversion from \eqref{eq:inte_1} to \eqref{eq:inte_2} can be found in \cite{kolter2022adversarial}. 

\subsection{Availability Adversarial Attack}

The integrity attack with the MILP reformulation has been well studied in the literature. However, it requires the attackers not only hijack the sensor measurements from the remote server, but also have the ability to manipulate their values and send them back to the operator. Compared to the integrity attack, the availability attack is much cheaper, as it only requires the attackers to block measurements \cite{pan2018cyber}. 

First, the temporal information, such as month, date and time, cannot be attacked as it can be easily obtained and verified by the operator. Therefore, only K1-K6 features are prone to attack (flexible feature, indexed by $\mathcal{I}_{flex}$) while the temporal feature is fixed (indexed by $\mathcal{I}_{fix}$). Second, once the K1-K6 features are blocked, the operator can have different choices to impute the missing value(s). In detail, defining an integer vector $\bm{m}\in\{0,1\}^{|\mathcal{I}_{flex}|}$ and vector $\bm{c}\in\mathbb{R}^{p}$, the imputed input data can be represented as $\bm{z}_1 = \text{diag}([\bm{m},\bm{1}])\cdot\bm{x} + \text{diag}([\bm{1} - \bm{m}, \bm{0}])\cdot\bm{c}$ where $\bm{1}$ and $\bm{0}$ are all one and zero vectors with proper dimensions\footnote{Without losing generality, we assume that the flexible features are concatenated in front of the fixed features.}. The imputed vector $\bm{c}$ reflects the operator's choice to compensate for the missing features. When $\bm{m}(i) = 0$, this feature is unavailable and $\bm{z}_1(i) = \bm{c}(i)$. For simplicity, $\bm{c}$ can be chosen as $\bm{0}$ or the average value $\frac{1}{N}\sum_n \bm{x}_n$ in this paper.

Similarly to \eqref{eq:inte_1}, the optimal availability adversarial attack can be obtained by solving the following optimization problem:
\begin{equation}\label{eq:avai_1}
\begin{split}
& {\min / \max}_{\bm{m}, \bm{z}_{1},\dots,\bm{z}_{d+1}}  \bm{z}_{d+1}  \\
\text{s.t.} \quad & \bm{z}_1 = \text{diag}([\bm{m},\bm{1}])\cdot\bm{x} + \text{diag}([\bm{1} - \bm{m}, \bm{0}])\cdot\bm{c} \\
& \bm{z}_{i+1} = \max{\{\bm{0}, \bm{W}_i\bm{z}_i + \bm{b}_i \}},\quad i = 1,\cdots,d-1 \\
& \bm{z}_{d+1} = \bm{W}_d \bm{z}_d + \bm{b}_d \\
& \bm{m} \in \{0,1\}^{|\mathcal{I}_{flex}|} \\
& |\mathcal{I}_{flex}| - \sum \bm{m} \leq \beta
\end{split}
\end{equation}
in which the last constraint represents the attacker's budget. For example, at most $\beta$ number of features can be blocked.

Using the same big-M technique in \eqref{eq:inte_2}, \eqref{eq:avai_1} can also be reformulated as MILP:
\begin{equation}\label{eq:avai_2}
\begin{split}
& {\min / \max}_{\bm{m}, \bm{z}_{1},\dots, \bm{z}_{d+1}, \bm{v}_1, \ldots, \bm{v}_{d-1}}  \bm{z}_{d+1} \\
\text{s.t.} \quad  & \bm{z}_{i+1} \geq \bm{W}_i \bm{z}_i + \bm{b}_i, \quad i=1,\dots,d-1 \\
& \bm{z}_{i+1} \geq 0, \quad i=1, \dots, d-1 \\
& \bm{u}_i \cdot \bm{v}_i \geq  \bm{z}_{i+1}, \quad i=1, \dots,d-1 \\
& \bm{W}_i \bm{z}_i + \bm{b}_i \geq \bm{z}_{i+1} + (\bm{1}-\bm{v}_i) \bm{l}_i, \quad i=1,\ldots,d-1 \\
& \bm{v}_i \in \{0,1\}^{|\bm{v}_i|}, \quad i=1, \ldots,d-1 \\
& \bm{z}_1 = \text{diag}([\bm{m},\bm{1}])\cdot\bm{x} + \text{diag}([\bm{1} - \bm{m}, \bm{0}])\cdot\bm{c} \\
& \bm{m} \in \{0,1\}^{|\mathcal{I}_{flex}|} \\
& |\mathcal{I}_{flex}| - \sum \bm{m} \leq \beta \\
& \bm{z}_{d+1} = \bm{W}_d \bm{z}_d + \bm{b}_d
\end{split}
\end{equation}

The user-defined layer output bounds $\bm{u}_i$ and $\bm{l}_i$ are essential to solve \eqref{eq:avai_1}. First, the solution of \eqref{eq:avai_2} may be sub-optimal to \eqref{eq:avai_1} if the bounds do not cover the output of layer $i$ at the optimal value of \eqref{eq:avai_1}. Second, if the bounds are too large, the MILP can be slow to converge. To inform proper bounds on the output of each linear layer, convex bound propagation can be applied as follows. First, define $\bm{l}_0 = \bm{u}_0 = \bm{x}$. Then the initial bounds on $\bm{z}_1$ can be defined as follows:
\begin{equation}
    \begin{split}
        \bm{l}_0[\mathcal{I}_{flex}] = \min{\{\bm{c}, \bm{x}\}} \\
        \bm{u}_0[\mathcal{I}_{flex}] = \max{\{\bm{c}, \bm{x}\}}
    \end{split}
\end{equation}

Considering \eqref{eq:nn_explicit}, the bounds on the output of each layer before the last can be propagated by
\begin{equation}
    \begin{split}
    \hat{\bm{l}}_i & = \max{\{\bm{0}, \bm{l}_i\}} \\
        \hat{\bm{u}}_i & = \max{\{\bm{0}, \bm{u}_i\}} \\
        \bm{l}_{i+1} & = \max{\{\bm{0}, \bm{W}_i\}}\cdot \hat{\bm{l}}_i + \min{\{\bm{0},\bm{W}_i\}}\cdot \hat{\bm{u}}_i + \bm{b}_i\\
        \bm{u}_{i+1} & = \min{\{\bm{0}, \bm{W}_i\}}\cdot \hat{\bm{l}}_i + \max{\{\bm{0},\bm{W}_i\}}\cdot \hat{\bm{u}}_i + \bm{b}_i
    \end{split}
\end{equation}
for $i=0,\cdots,d-2$. By the above formulation, there should be at least two layers in the network. Otherwise, the ANN will become a linear regression.

\begin{remark}
    A small value can be subtracted and added on $\bm{l}_0[\mathcal{I}_{flex}]$ and $\bm{u}_0[\mathcal{I}_{flex}]$ respectively to avoid numerical instability when solving MILP.
\end{remark}

\section{Adversarial Training against Availability Attack}

To overcome the availability attack defined in \eqref{eq:avai_1} and \eqref{eq:avai_2}, adversarial training can be implemented in which the loss of the worst-case attack is minimized in each epoch. Defining $\bm{z}_n = \mathcal{M}(\bm{m}) = \text{diag}([\bm{m},\bm{1}])\cdot\bm{x}_n + \text{diag}([\bm{1} - \bm{m}, \bm{0}])\cdot\bm{c}_n$, the adversarial loss function is written as
\begin{equation}\label{eq:loss_adver}
    \mathcal{L}^{adv}(\bm{\theta}) = \max_{\bm{m}}\frac{1}{N}\sum_{n=1}^N\left(y_n - f(\bm{z}_n;\bm{\theta})\right)^2
\end{equation}

Similarly to SGD in clean training \eqref{eq:sgd_plain}, \eqref{eq:loss_adver} can be minimized iteratively as
\begin{equation}\label{eq:sgd_adver}
    \bm{\theta}:=\bm{\theta} - \frac{\alpha}{|\bm{B}_n|} \sum_{(\bm{x},\bm{y})\in\bm{B}_n}\nabla_{\bm{\theta}}\max_{\bm{m}}(y-f(\bm{z};\bm{\theta}))^2
\end{equation}

According to Danskin's Theorem \cite{madry2017towards}, the gradient of the maximization problem equals the gradient of the loss function at the maxima. Therefore,
\begin{equation}
    \nabla_{\bm{\theta}}\max_{\bm{m}}(y-f(\bm{z};\bm{\theta}))^2 = \nabla_{\bm{\theta}}(y - f(\bm{z}^*;\bm{\theta}))^2
\end{equation}
where $\bm{z}^*$ is calculated from the optimal solution $\bm{m}^*$ of the adversarial loss \eqref{eq:loss_adver}. Consequently, the loss function of adversarial training for availability attacks can be written as
\begin{equation}\label{eq:loss_avai}
    \mathcal{L}^{avai}(\bm{\theta}) = \mathcal{L}(\bm{\theta}) + \beta_{max}\mathcal{L}^{adv}_{max}(\bm{\theta}) + \beta_{min}\mathcal{L}^{adv}_{min}(\bm{\theta})
\end{equation}

In \eqref{eq:loss_avai}, $\mathcal{L}(\cdot)$, $\mathcal{L}^{adv}_{max}(\cdot)$, and $\mathcal{L}^{adv}_{min}(\cdot)$ represent the clean loss, maximization, and minimization of adversarial loss \eqref{eq:loss_adver} respectively. Instead of solely training on the adversarial instance, the clean loss is also added to improve the accuracy on clean dataset \cite{zhang2019theoretically}. 

\begin{remark}
Although the proposed availability adversarial attack and adversarial training are based on feedforward neural networks, they can also be extended to other types of piecewise linear layer, such as the convolutional layer. Due to the page limit, the discussion on convolutional neural networks will be left for future work.
\end{remark}

\section{Simulation and Result}

The optimal integrity and availability adversarial attacks are constructed using CVXPY \cite{diamond2016cvxpy} with Gurobi solver. Python muti-processing is used to accelerate the MILP optimization in \eqref{eq:inte_2} and \eqref{eq:avai_2}. The deep load forecasting model is trained using PyTorch on an RTX 3090 graphic card. The outputs of the layers in ANN are 40-20-10-1. Adam optimizer is used with initial learning rate of 0.0005 and cosine annealing. For the original dataset \cite{farrokhabadi2022day}, outliers beyond three-sigma are removed and we randomly separate the dataset into train and test with proportion 8:2. Flexible features are scaled into [0,1]. We train the model for 150 epochs and store the weights with best test set performance. For adversarial training, the hyperparameters are set to $\beta_{max}=\beta_{min} = 1$. The accuracy of the load forecasting model is reported by mean absolute percentage error (MAPE). For clean samples, we measure the distance between forecasted and ground-truth loads:
\begin{equation}
    \text{MAPE} = \frac{1}{N} \sum_{n=1}^N\frac{|y_{pred}^{clean, n} - y_{true}^n|}{y_{true}^n} \times 100\%
\end{equation}

For adversarial samples, we use the mean percentage error (MPE) to evaluate their output against the output of the clean sample:
\begin{equation}
    \text{MPE} = \frac{1}{N} \sum_{n=1}^N\frac{y_{pred}^{adv, n} - y_{pred}^{clean,n}}{y_{pred}^{clean,n}} \times 100\%
\end{equation}
as the adversarial attack is formulated on the trained neural network. 

In the following discussion, the model trained on clean dataset is referred as Clean Model, while the model trained through adversarial training is referred as Adver Model. Furthermore, to distinguish different attack attempts, we use AVAI(mode, $\bm{c}$, $\beta$) to represent the availability adversarial attack with mode$\in$\{max,min\} and $\bm{c}\in\{0,\text{mean}\}$. Similarly, INTE(mode, $\epsilon$) represents the integrity adversarial attack with attack strength $\epsilon$.

\subsection{Performance of Availability Adversarial Attack}

\begin{figure}[h]
     \centering
     \begin{subfigure}[b]{0.24\textwidth}
         \centering
         \includegraphics[width=\textwidth]{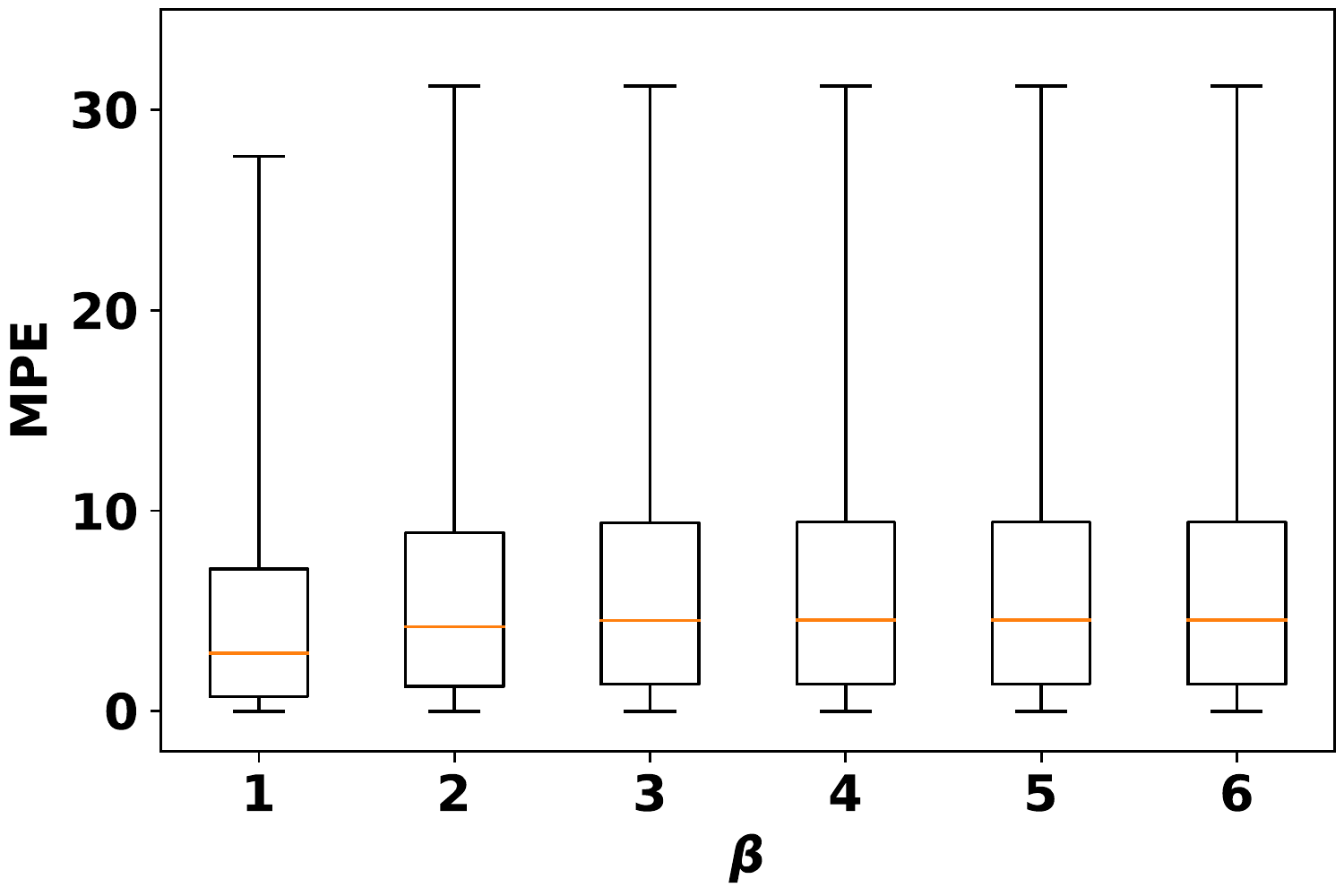}
         \caption{AVAI(max, 0, $\beta$)}
     \end{subfigure}
     \hfill
     \begin{subfigure}[b]{0.24\textwidth}
         \centering
         \includegraphics[width=\textwidth]{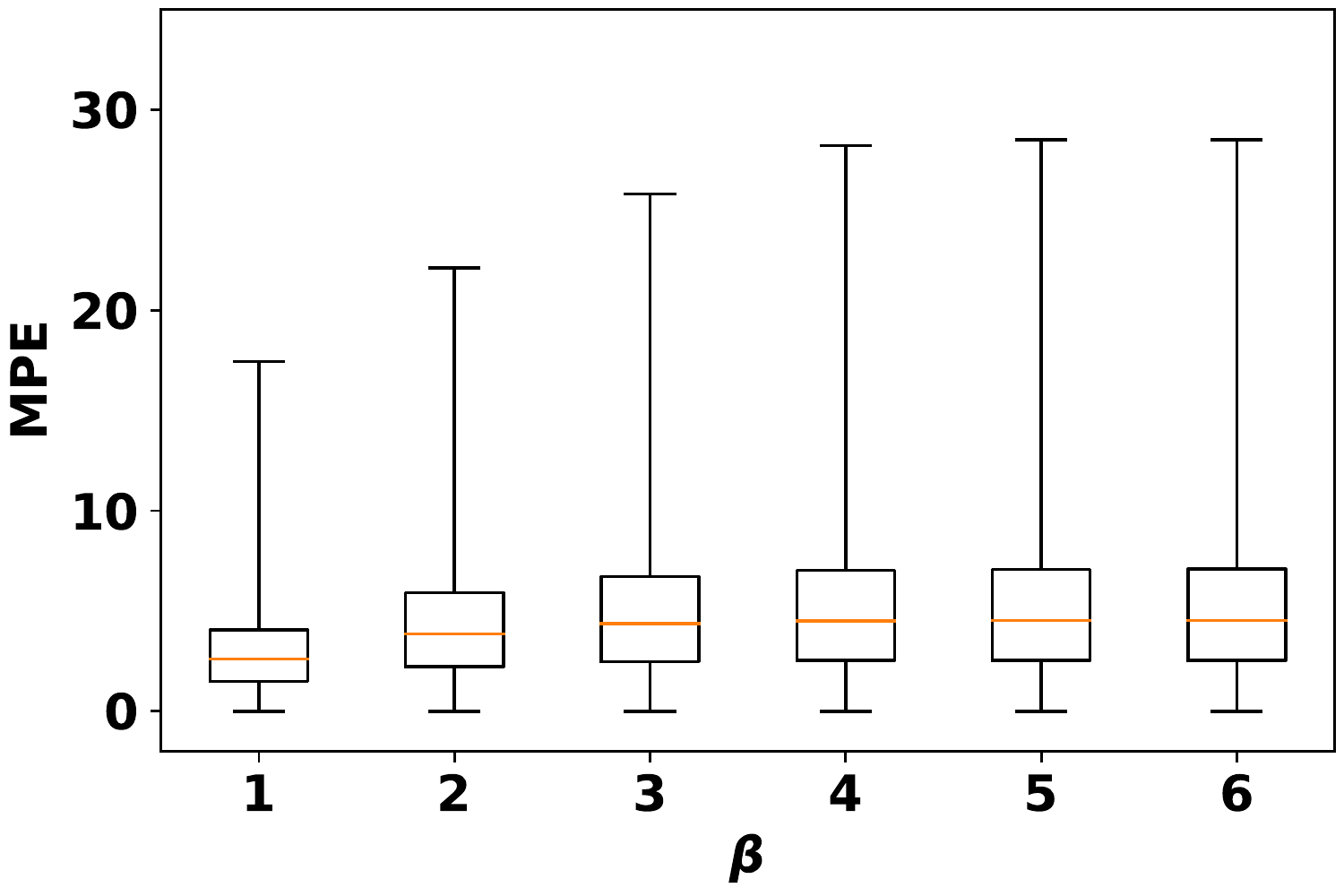}
         \caption{AVAI(max, mean, $\beta$)}
     \end{subfigure}
     \hfill
     \begin{subfigure}[b]{0.24\textwidth}
         \centering
         \includegraphics[width=\textwidth]{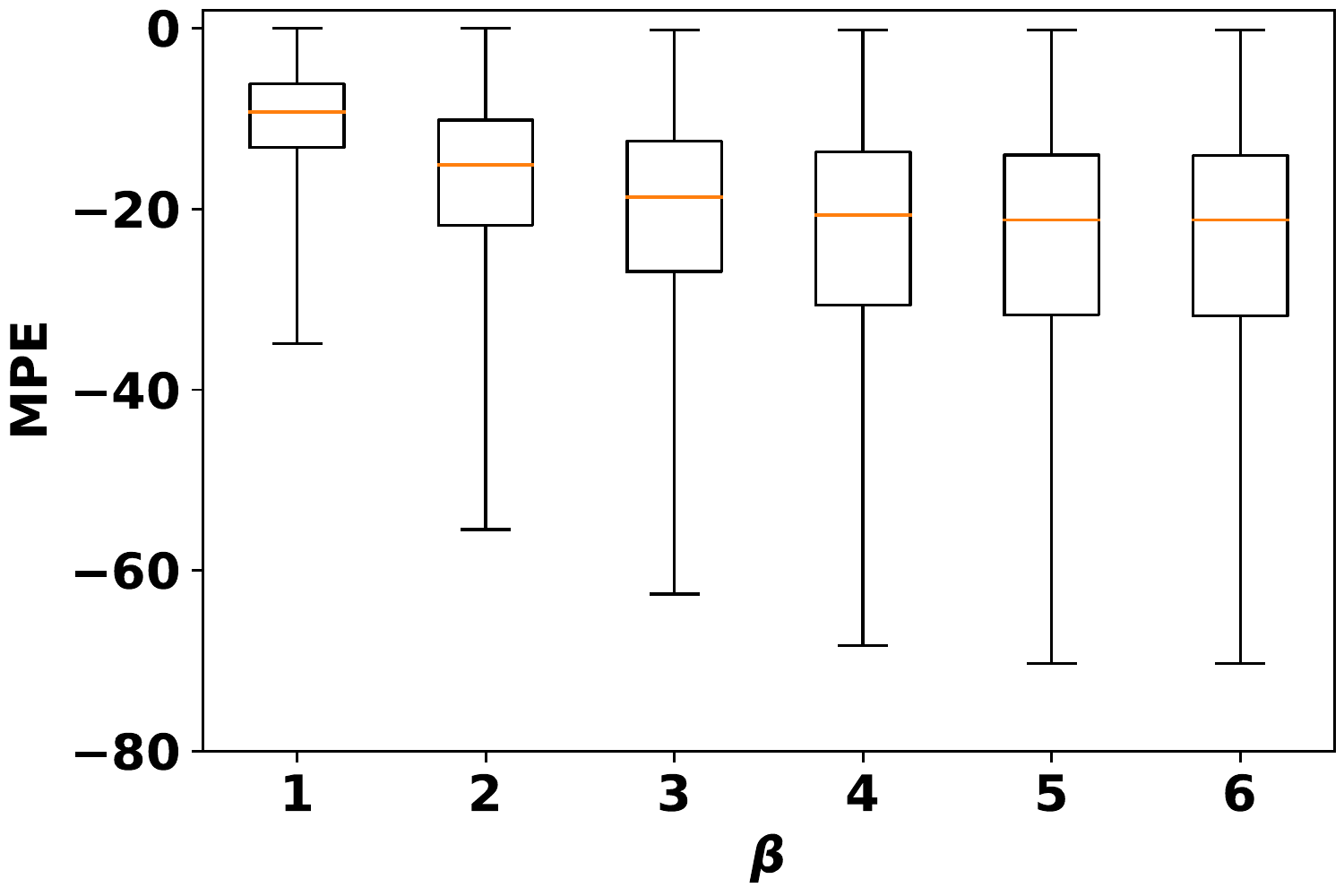}
         \caption{AVAI(min, 0, $\beta$)}
     \end{subfigure}
    \hfill
     \begin{subfigure}[b]{0.24\textwidth}
         \centering
         \includegraphics[width=\textwidth]{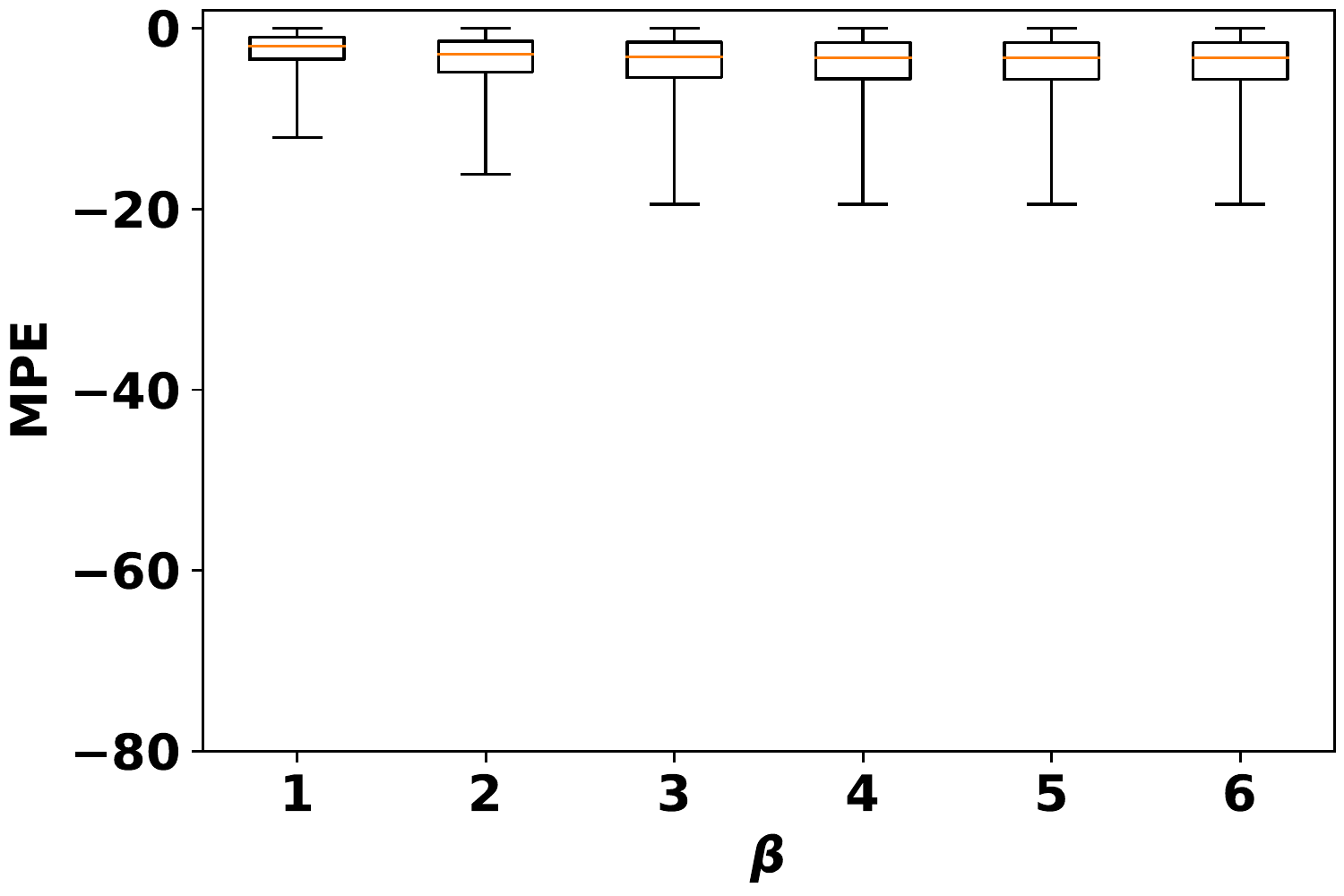}
         \caption{AVAI(min, mean, $\beta$)}
     \end{subfigure}
        \caption{MPEs on the clean model under availability adversarial attacks.}
        \label{fig:avai_clean}
\end{figure}

Fig. \ref{fig:avai_clean} compares the MPE deviation on the clean model under the availability adversarial attacks. In each of the box plot, the box represents the inter-quartile range (IQR) extending from the first quartile to the third of the data. The median of the data is indicated by the orange line. The whiskers cover the whole range of data. First, increasing the attack budget $\beta$ can increase the output deviations. However, this trend is not significant when $\beta \geq 3$. The output deviations with zero imputation $\bm{c}=0$ is more intense than average imputation as no feature information can be extracted. Second, different data can have different vulnerabilities on the availability attack. For example, some samples can maintain the output with MPE=0 while others are altered significantly.

\begin{figure}[h]
     \centering
     \begin{subfigure}[b]{0.24\textwidth}
         \centering
         \includegraphics[width=\textwidth]{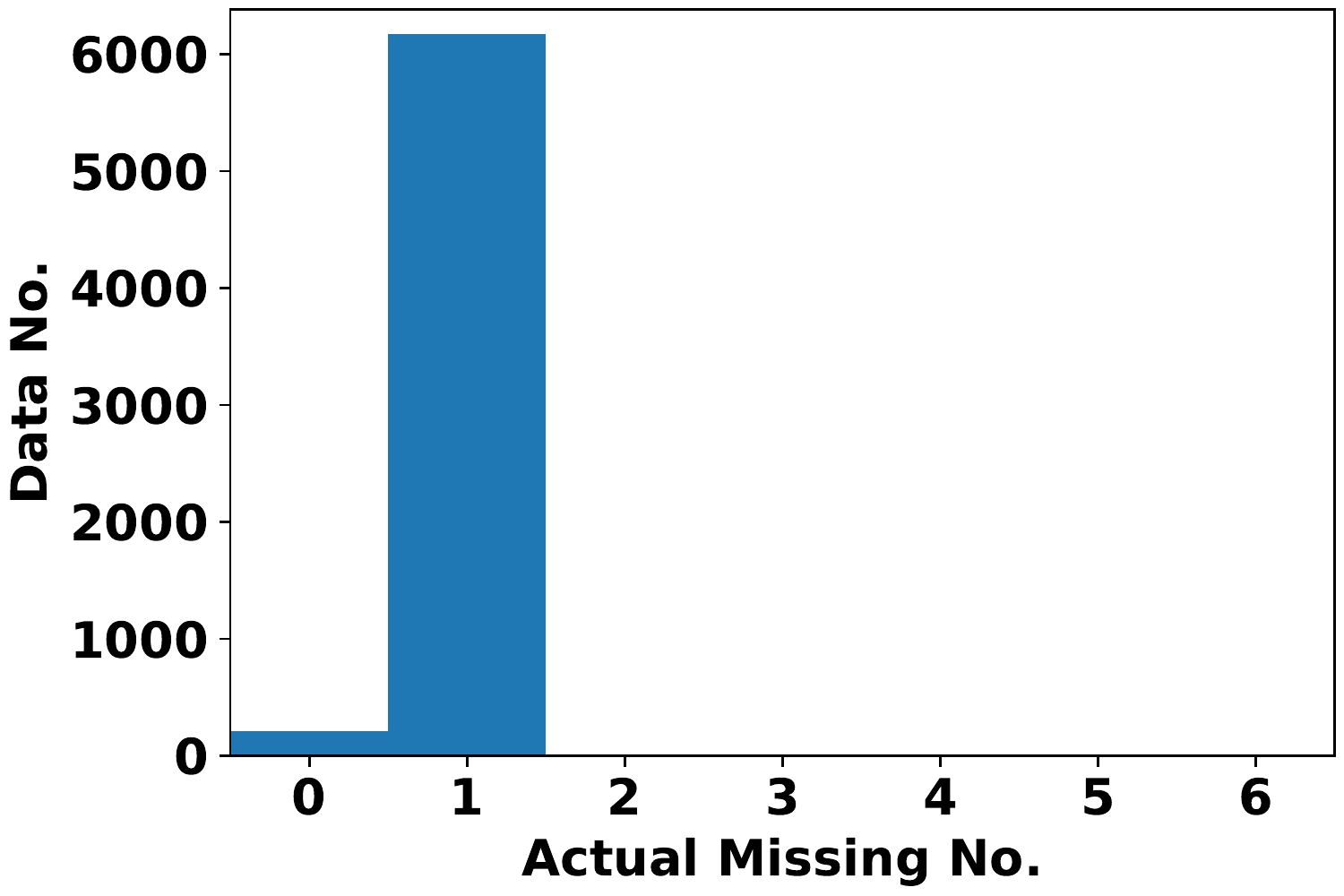}
         \caption{$\beta=1$}
     \end{subfigure}
     \hfill
     \begin{subfigure}[b]{0.24\textwidth}
         \centering
         \includegraphics[width=\textwidth]{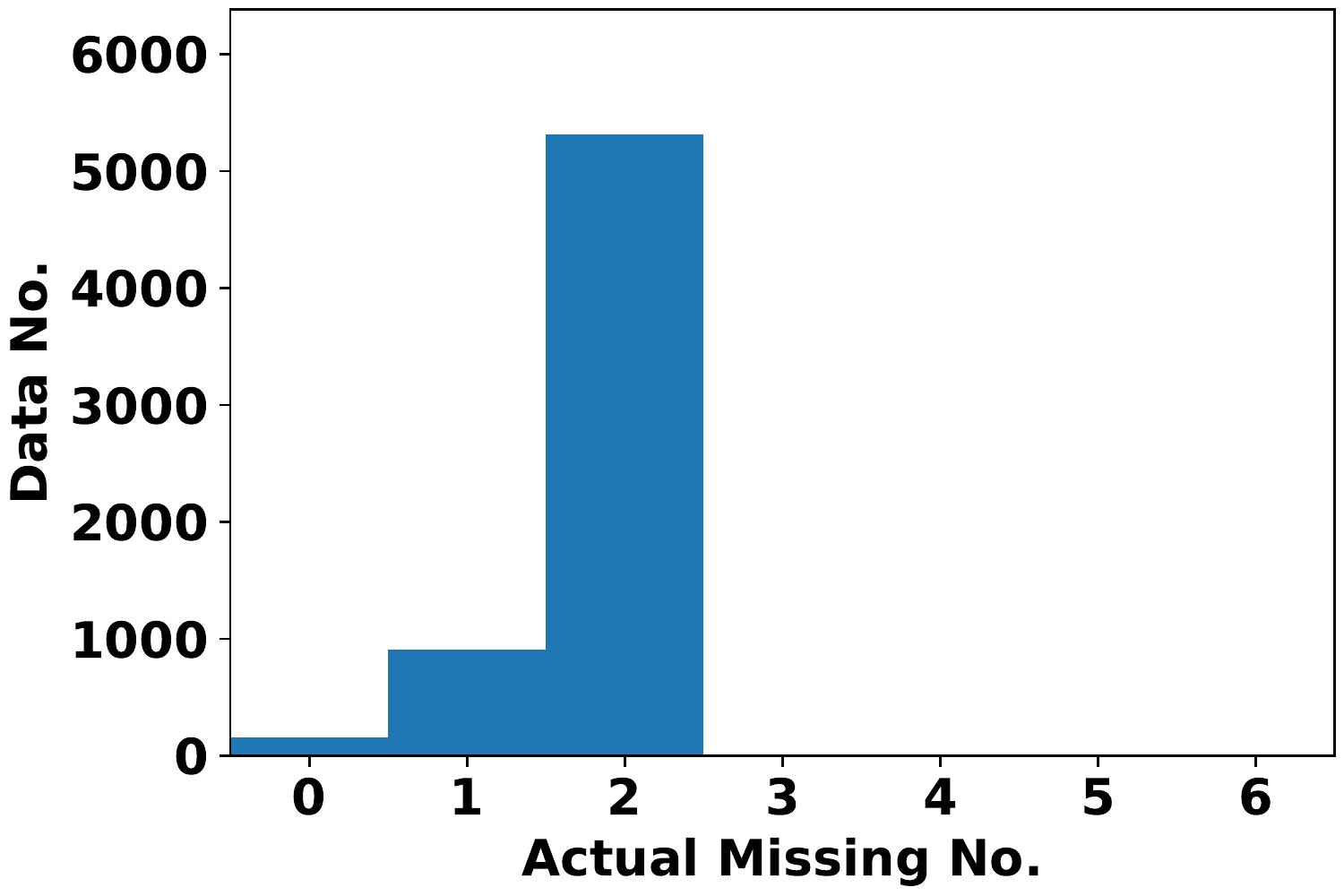}
         \caption{$\beta=2$}
     \end{subfigure}
     \hfill
     \begin{subfigure}[b]{0.24\textwidth}
         \centering
         \includegraphics[width=\textwidth]{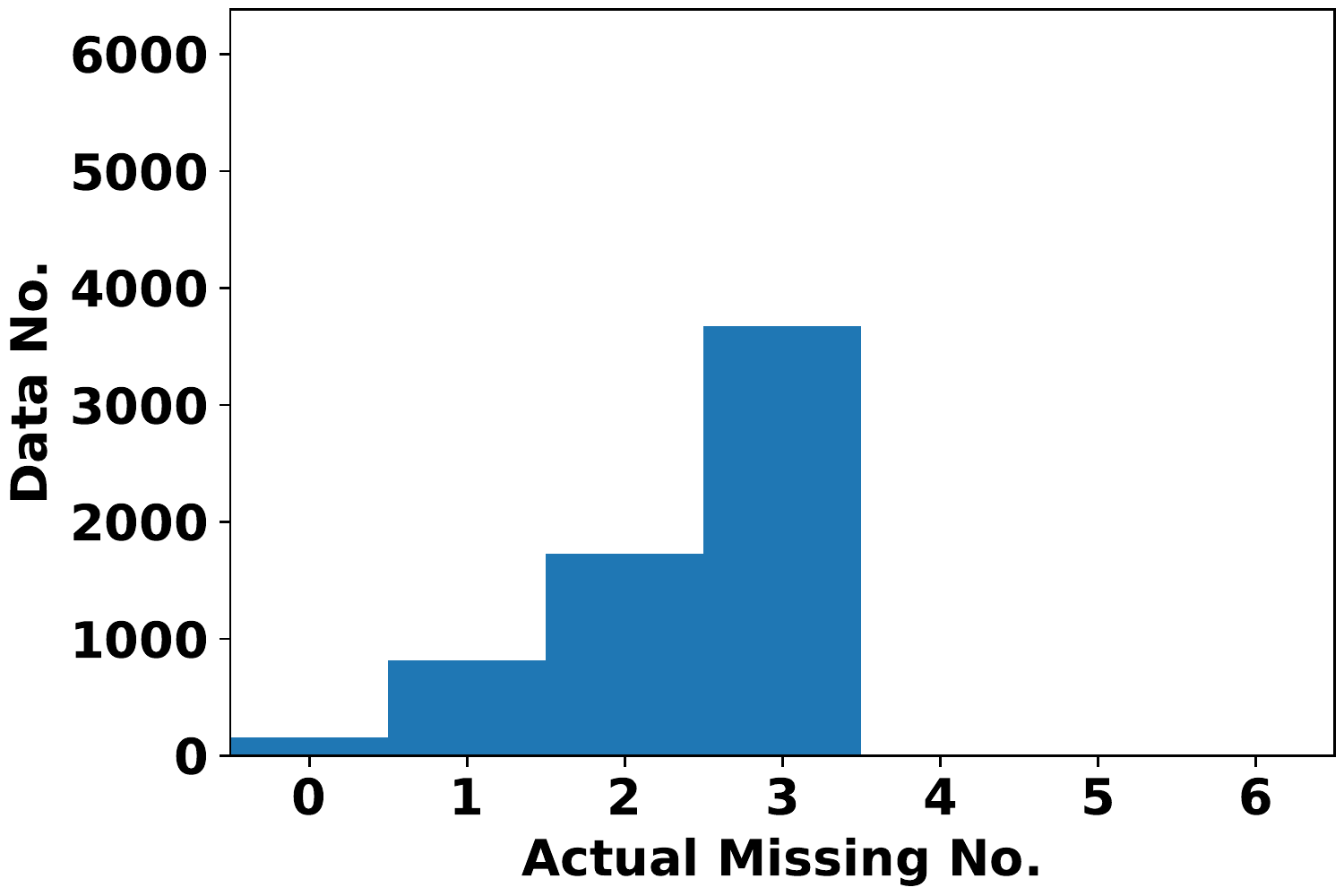}
         \caption{$\beta=3$}
     \end{subfigure}
    \hfill
     \begin{subfigure}[b]{0.24\textwidth}
         \centering
         \includegraphics[width=\textwidth]{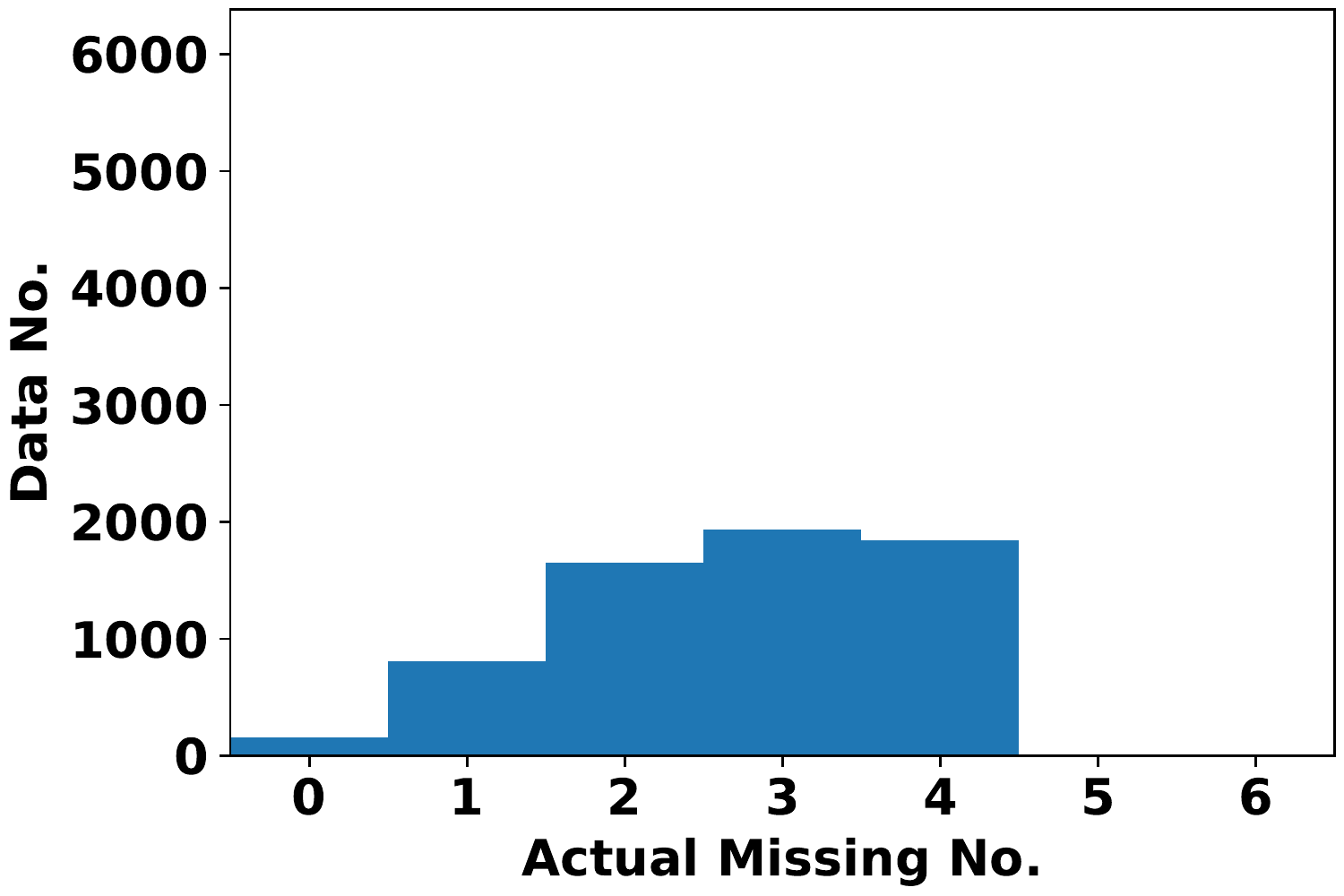}
         \caption{$\beta=4$}
     \end{subfigure}
     \hfill
     \begin{subfigure}[b]{0.24\textwidth}
         \centering
         \includegraphics[width=\textwidth]{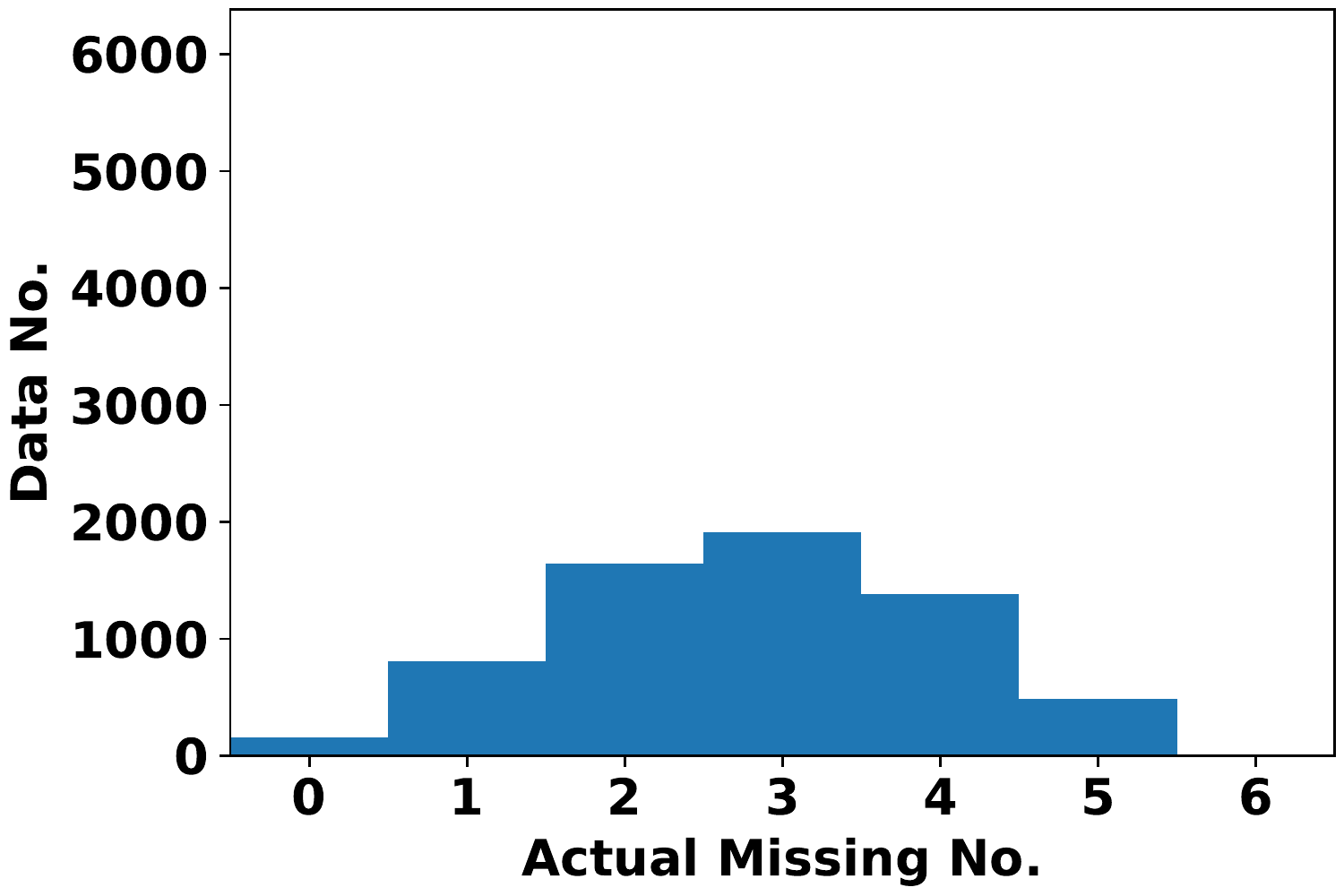}
         \caption{$\beta=5$}
     \end{subfigure}
     \hfill
     \begin{subfigure}[b]{0.24\textwidth}
         \centering
         \includegraphics[width=\textwidth]{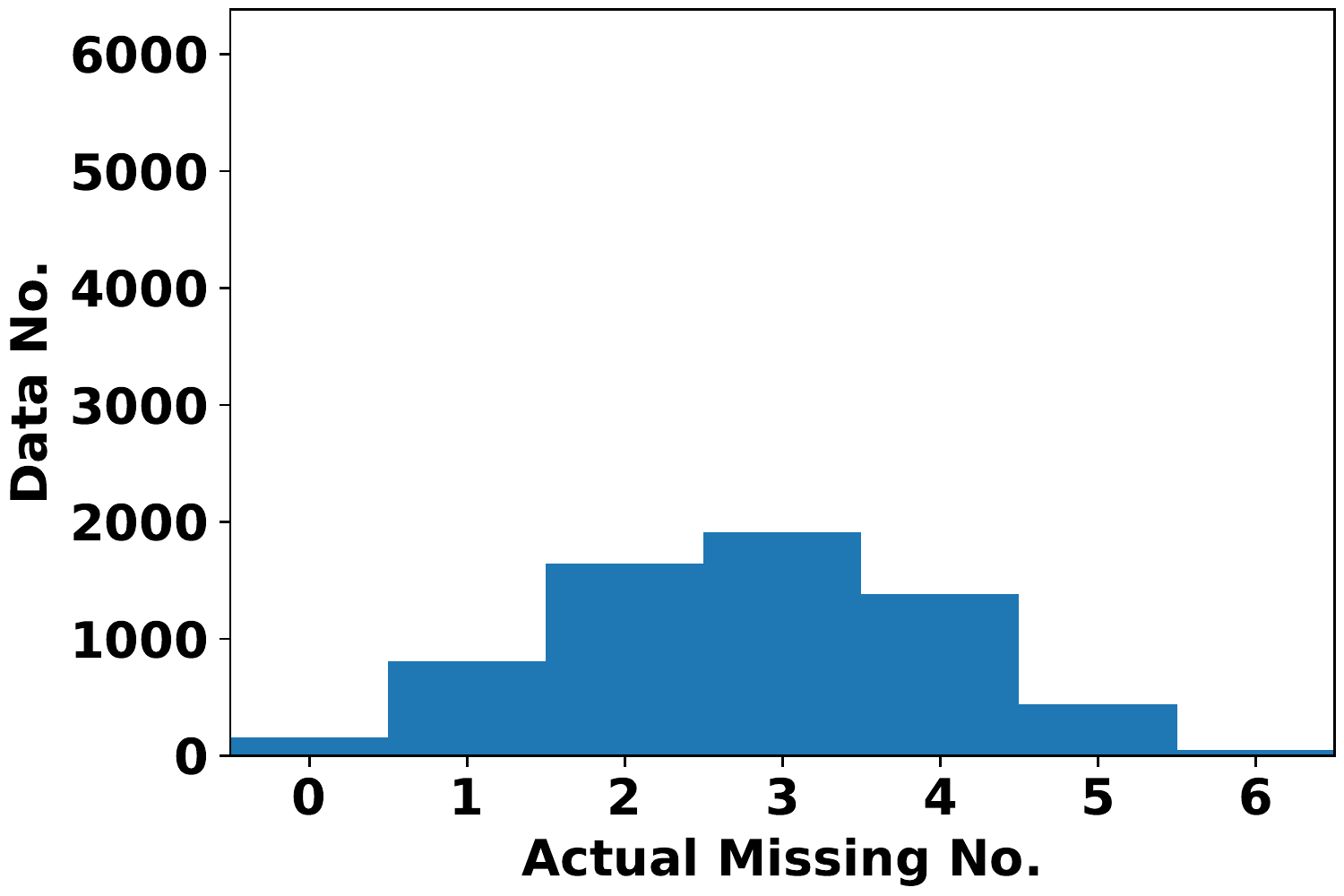}
         \caption{$\beta=6$}
     \end{subfigure}
        \caption{Actual missing numbers under different attack budget $\beta$. The AVAI(min, mean, $\beta$) attack strategy is used.}
        \label{fig:avai_clean_missing}
\end{figure}

To better see the dependency of attack performance on the attack budget $\beta$, Fig. \ref{fig:avai_clean_missing} records the number of actual missing features under AVAI(min, mean, $\beta$) attack (corresponding to Fig. 1(d)). First, it is clearly shown that a small proportion of data cannot be attacked regardless of the choices of $\beta$, which results in MPE=0. These data points are least sensitive to AVAI(min, mean, $\beta$) attack, as blocking any subset of their features can only increase the load forecast. Second, most of the actual missing numbers locate at 3 even when $\beta=6$. This further implies that in most cases, blocking more features may not result in a stronger attack, although increasing $\beta$ can have more flexibility to allocate the attack position. 

\begin{figure}[h]
     \centering
     \begin{subfigure}[b]{0.24\textwidth}
         \centering
         \includegraphics[width=\textwidth]{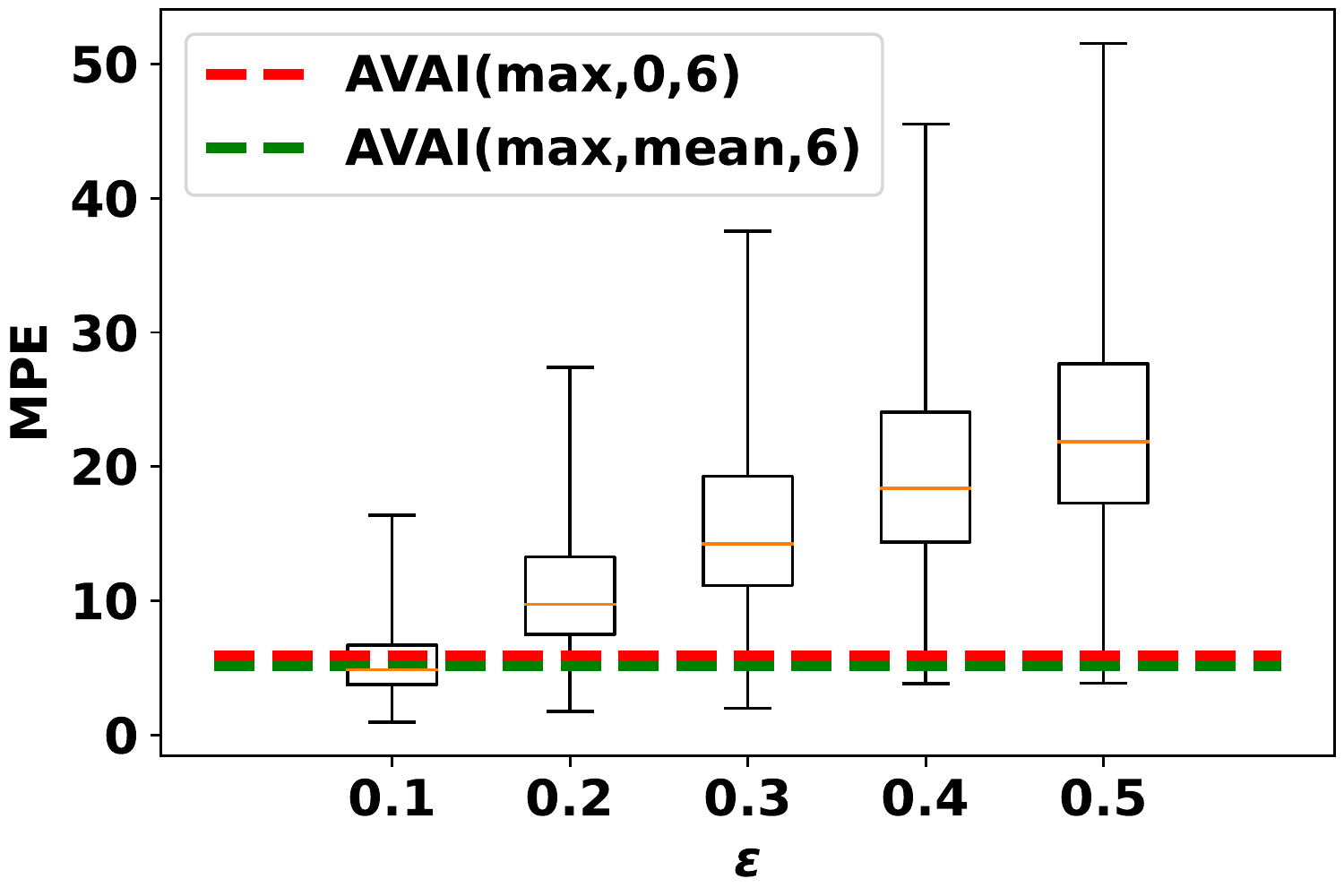}
         \caption{INTE(max, $\epsilon$)}
     \end{subfigure}
     \hfill
     \begin{subfigure}[b]{0.24\textwidth}
         \centering
         \includegraphics[width=\textwidth]{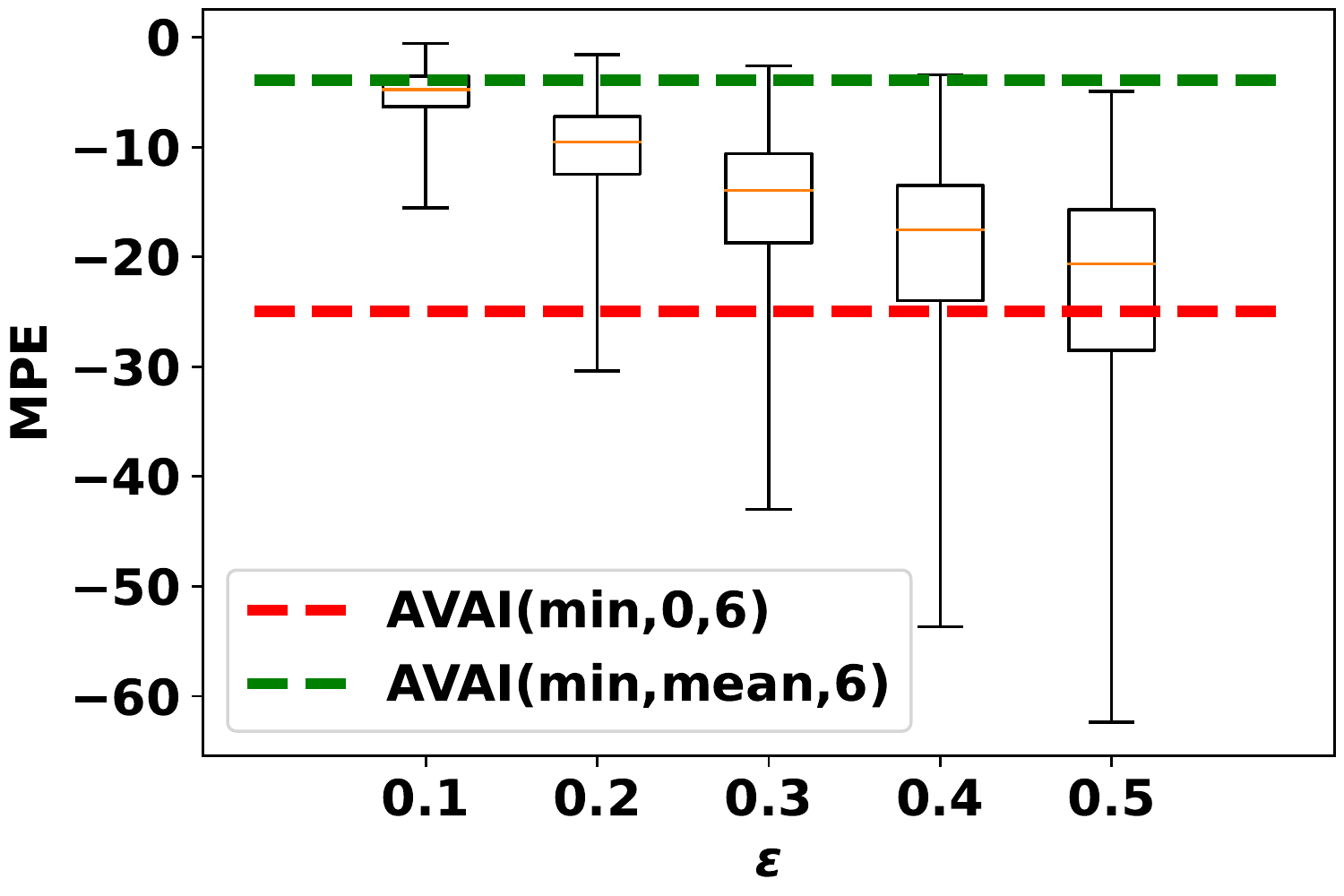}
         \caption{INTE(min, $\epsilon$)}
     \end{subfigure}
        \caption{MPE on the clean model under integrity adversarial attacks.}
        \label{fig:inte_clean}
\end{figure}

Fig. \ref{fig:inte_clean} records the MPE of integrity adversarial attacks solved by MILP \eqref{eq:inte_2}. The median MPEs of availability attacks with $\beta=6$ are also plotted as a reference. When maximizing the load forecast (Fig. \ref{fig:inte_clean}(a)), availability attacks can give output deviations comparable to integrity attacks when $\epsilon$ is small, e.g. with $\epsilon=0.1$. However, when $\epsilon$ increases, the MPE of the integrity attack is much higher than the availability attacks. When minimizing the load forecast (Fig. \ref{fig:inte_clean}(b)), AVAI(min, mean, 6) can persistently outperform integrity attacks. Since the input features are scaled into $[0,1]$, it is not realistic to have $\epsilon$ larger than 0.2. Meanwhile, as it is is much cheaper than the integrity attack, the availability attack is a promising attack strategy for the attacker. However, how to balance the cost and impact of this attack is unsolved and we will leave it for future work.

Solving the MILP can be time consuming, especially when the number of integer variables is large. The computational time of both availability attacks are summarized in Table \ref{tab:time}. As adversarial training requires solving the optimal availability attacks, faster computation is beneficial to efficient training. Recall that the size of the dataset is 32k, which requires half hour to train single epoch. After using parallel computation, the computational time is significantly reduced to 1.7 min/epoch, which is acceptable for a real-time application.

\begin{table}[h]
    \centering
    \footnotesize
    \caption{Average Computational Time (ms/sample)}
    \begin{tabular}{r|c}\hline
        \textbf{Attack Types} & \textbf{Time} \\\hline\hline
        \textbf{AVAI-Parallel} & 3.16 \\\hline
        \textbf{AVAI-Sequential} & 49.73 \\\hline
    \end{tabular}
    \label{tab:time}
\end{table}

\subsection{Performance of Adversarial Training}

Adversarial training is implemented in this section. Since there are two imputation strategies discussed in this paper, the performances are reported on both models trained adversarially with $\bm{c}=0$ and $\bm{c}=\text{mean}$. In addition, the attack budget is set to $\beta=6$ in adversarial training.

First, the MAPEs on the clean samples for both clean model and adversarial model are summarized in Table \ref{tab:mape}. Adversarial training can inevitably deteriorate the performance of the model in clean samples by 1.0\%. As $\bm{c}=0$ is a stronger attack attempt than $\bm{c}=\text{mean}$, model trained on AVAI(mode, 0, 6) has higher MAPE than it trained on AVAI(mode, mean, 6).

\begin{figure}[h]
     \centering
     \begin{subfigure}[b]{0.24\textwidth}
         \centering
         \includegraphics[width=\textwidth]{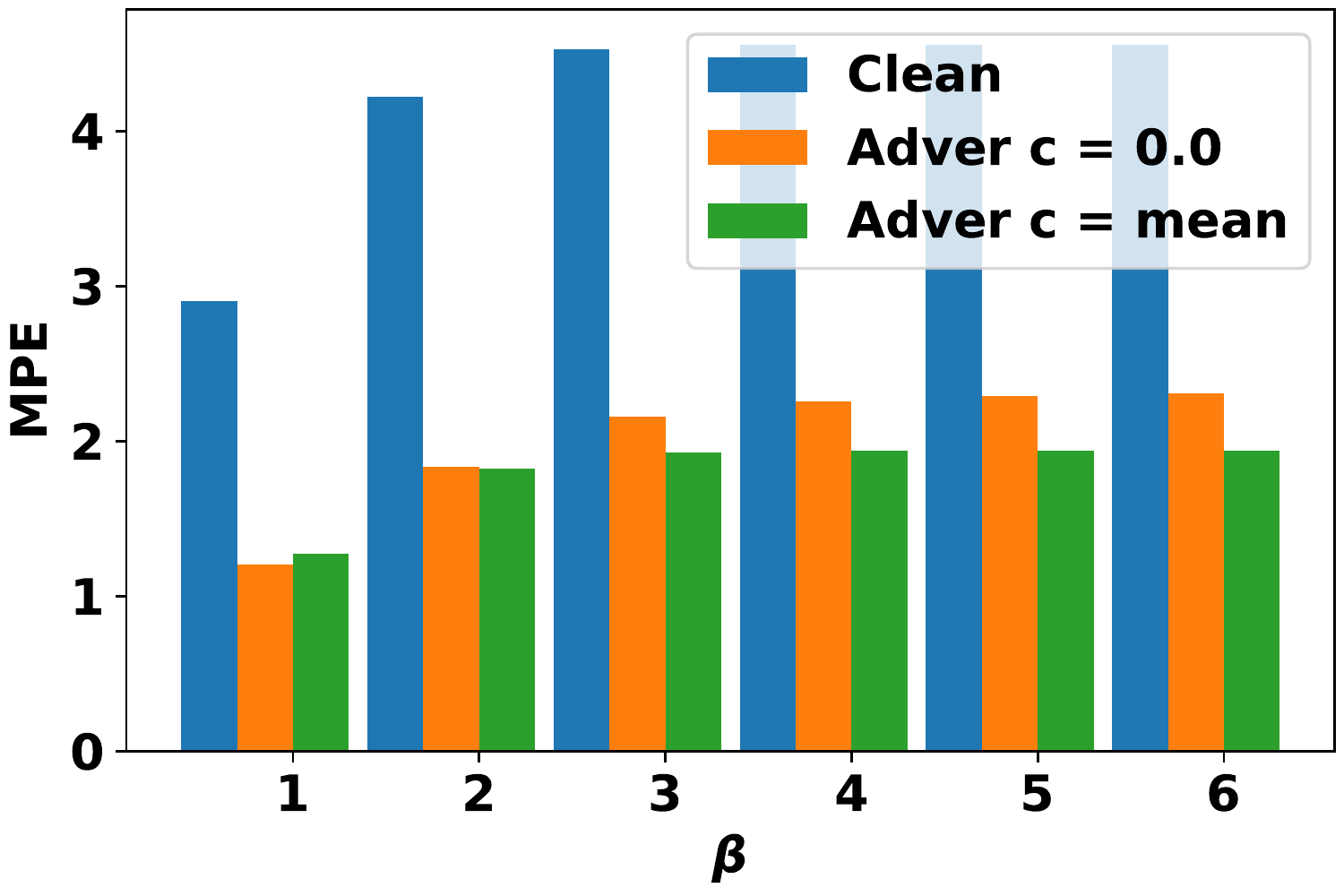}
         \caption{AVAI(max, 0, $\beta$)}
     \end{subfigure}
     \hfill
     \begin{subfigure}[b]{0.24\textwidth}
         \centering
         \includegraphics[width=\textwidth]{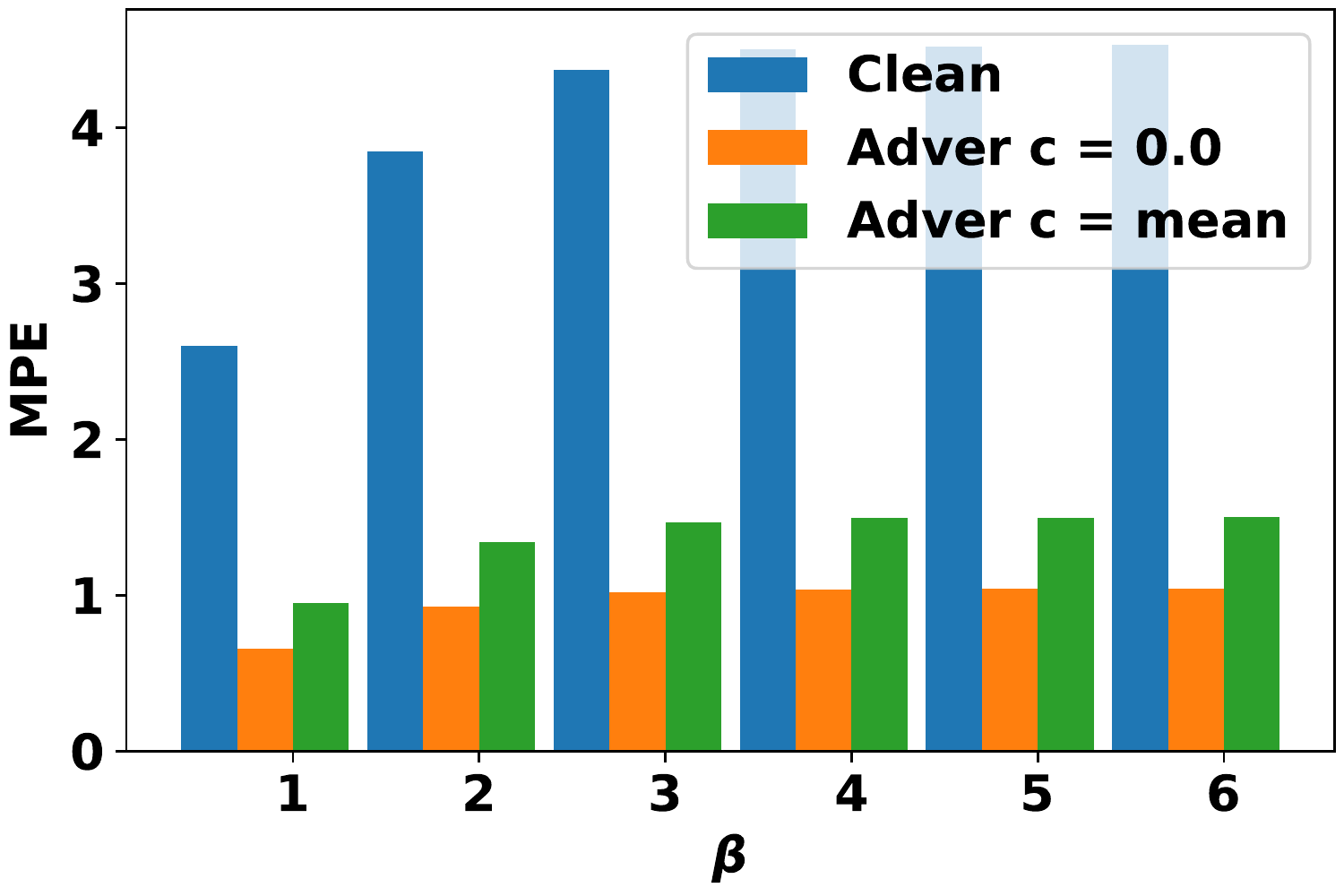}
         \caption{AVAI(max, mean, $\beta$)}
     \end{subfigure}
     \hfill
     \begin{subfigure}[b]{0.24\textwidth}
         \centering
         \includegraphics[width=\textwidth]{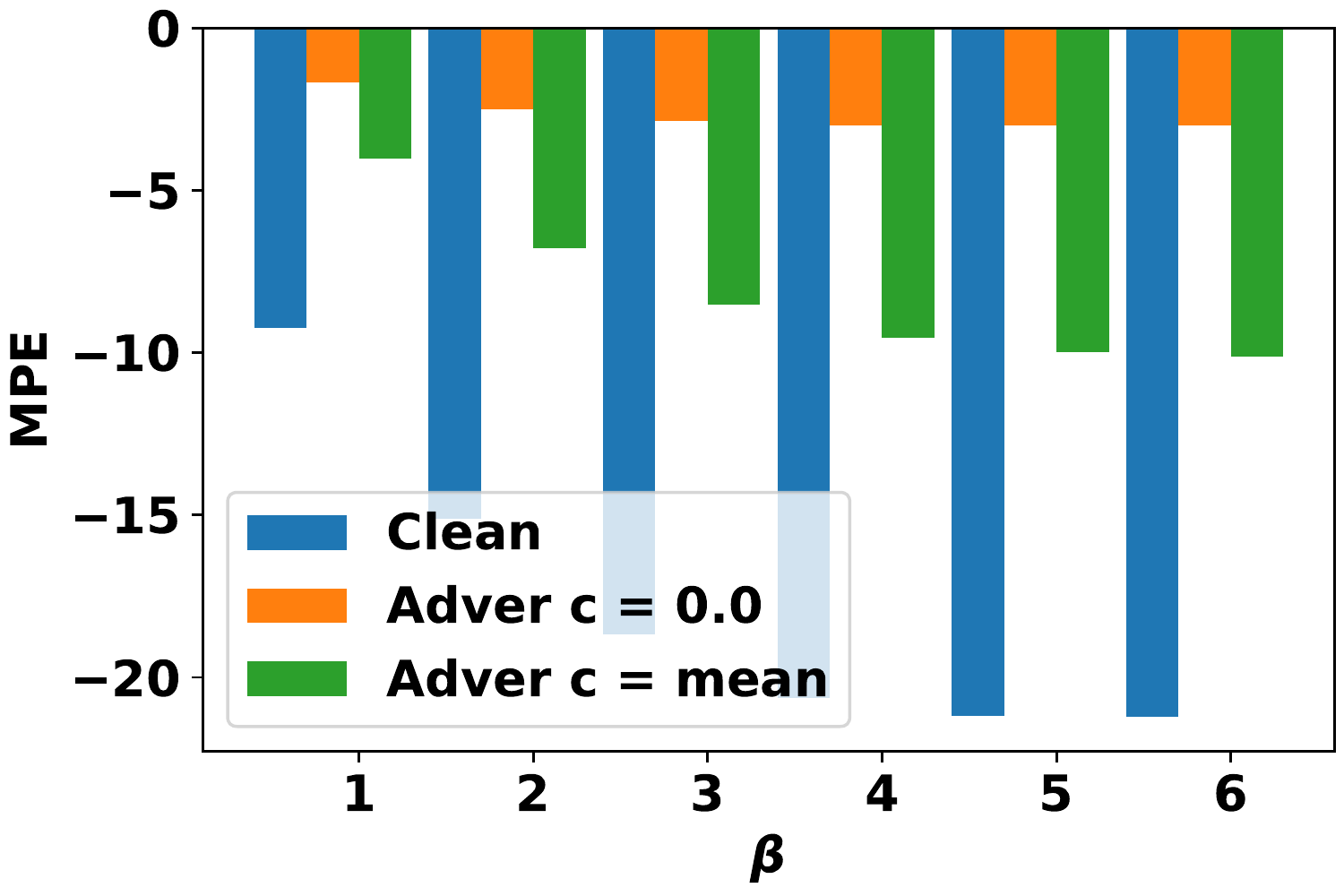}
         \caption{AVAI(min, 0, $\beta$)}
     \end{subfigure}
    \hfill
     \begin{subfigure}[b]{0.24\textwidth}
         \centering
         \includegraphics[width=\textwidth]{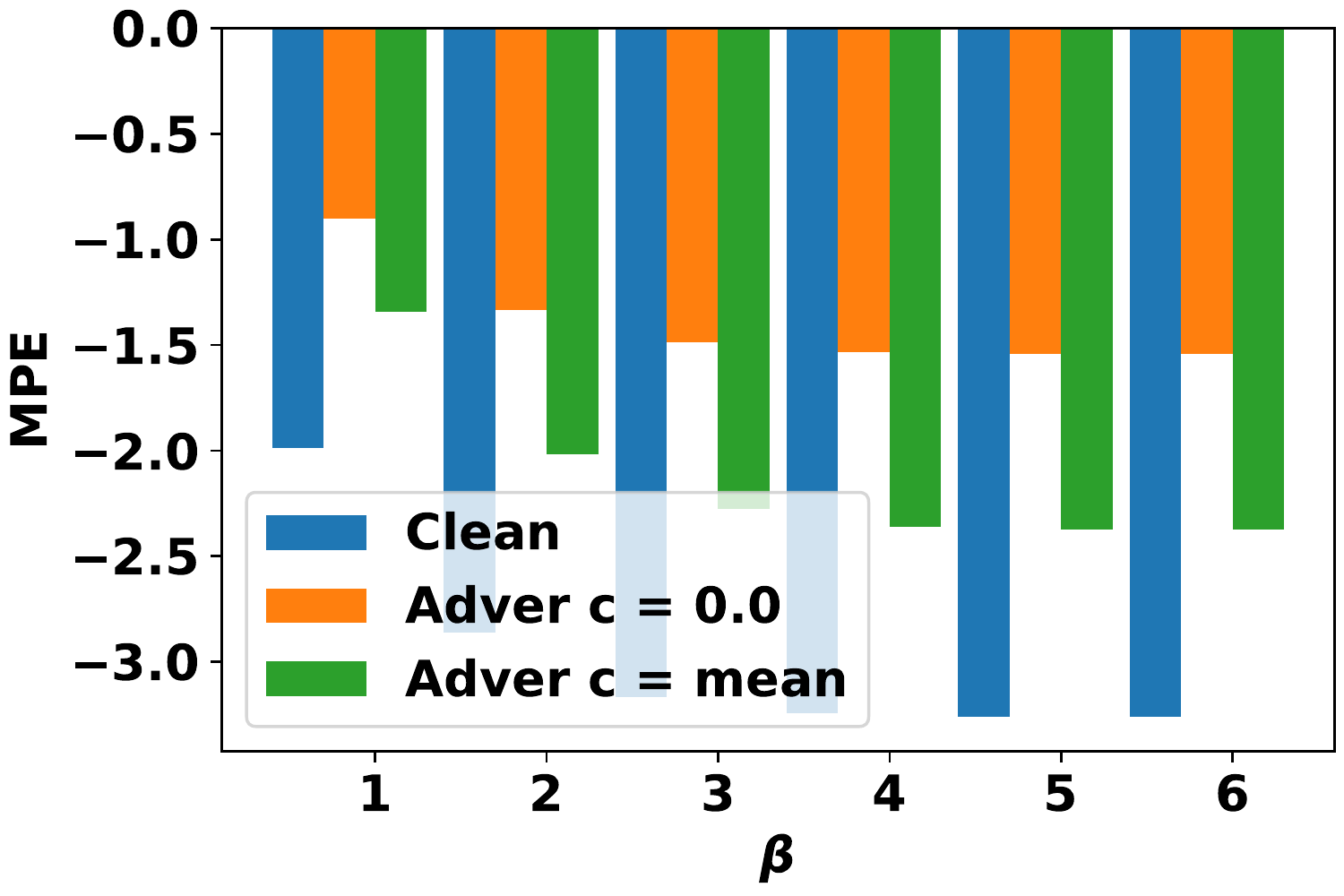}
         \caption{AVAI(min, mean, $\beta$)}
     \end{subfigure}
        \caption{Performances of adversarial training on availability attacks. The medians are taken for all samples in test dataset.}
        \label{fig:adver}
\end{figure}

Second, the MPE on the adversarially attacked samples under different training situations are summarized in Fig. \ref{fig:adver}. After the adversarial training, the MPEs are reduced by more than 50\% in general. The stronger adversarial training situation with $\bm{c} = 0$ can have better robustness, but sacrifices more in clean accuracy (Table \ref{tab:mape}). It is interesting to observe that the model trained with $\bm{c} = 0$ can also have better performance on AVAI(mode, mean, 6) than the model trained directly with $\bm{c}=\text{mean}$. Moreover, the performance of model trained with $\bm{c} = 0$ has slightly higher MPEs than the model trained with $\bm{c} = \text{mean}$. Referring to \eqref{eq:loss_avai}, both AVAI(max, $\bm{c}$, 6) and AVAI(min, $\bm{c}$, 6) attacks contribute equally to the adversarial loss function. Therefore, the gradient descent tries to balance them during the training, although the AVAI(min, 0, 6) is much stronger than AVAI(max, 0, 6). To solve the problem, the hyperparameters can be set to $\beta_{max} > 1 > \beta_{min} > 0$ in \eqref{eq:loss_avai}.

\begin{table}[h]
    \centering
    \footnotesize
    \caption{MAPE on clean samples (in \%).}
    \begin{tabular}{r|c|c|c}\hline
                       & \textbf{Clean} & \textbf{Adver (c=0)} & \textbf{Adver (c=mean)} \\\hline\hline
        \textbf{Train} & 5.81 & 6.83 & 6.76 \\\hline
        \textbf{Test} & 6.07 & 6.95 & 6.88 \\\hline
    \end{tabular}
    \label{tab:mape}
\end{table}

\section{Conclusion}

This paper proposes a new availability adversarial attack on load forecasting model constructed by piece-wise linear neural network. The attack is optimally found through MILP subject to certain attack budget, and a countermeasure is given through adversarial training. The simulation results show that the availability attack can achieve attack performance comparable to that of the integrity counterpart. Meanwhile, adversarial training can effectively reduce MPE by more than 50\% on the adversarial samples while only increasing 1\% MAPE on the clean samples.

\bibliographystyle{IEEEtran}
\bibliography{IEEEabrv,Reference.bib}

\end{document}